\documentclass[letterpaper, 10 pt, conference]{ieeeconf}  

\IEEEoverridecommandlockouts                              

\usepackage{graphics} 
\usepackage{epsfig} 
\usepackage{mathptmx} 
\usepackage{times} 
\usepackage{amsmath} 
\usepackage{amssymb}  

\usepackage{graphicx}
\usepackage{amssymb}
\usepackage{amsmath}
\usepackage{multicol}
\usepackage{multirow}

\usepackage{graphicx}
\usepackage{amsmath}
\usepackage{subfigure}
\usepackage{multirow}
\usepackage{makecell} 
\usepackage{pifont}
\usepackage{graphicx}
\usepackage{amsmath}
\usepackage{amssymb}
\usepackage{booktabs}
\usepackage{amssymb}
\usepackage{colortbl}
\usepackage{graphicx}
\usepackage{multirow}
\usepackage{color}
\usepackage{bbding}
\usepackage{hyperref}

\title{\LARGE \bf
Sim2Real within 5 Minutes: Efficient Domain Transfer with Stylized Gaussian Splatting for Endoscopic Images
}

\author{Junyang Wu$^{1}$, Yun Gu$^{1,*}$ and Guang-Zhong Yang$^{1,*}$
\thanks{$^{1}$Junyang Wu, Yun Gu and Guang-Zhong Yang are with the Institute of Medical Robotics, Shanghai Jiao Tong University, Shanghai, 200240, China. }
\thanks{Corresponding Authors: Yun Gu~(yungu@ieee.org), Guang-Zhong Yang~(gzyang@sjtu.edu.cn) }
\thanks{This work is supported by Natural and Science Foundation of China (No.62373243). }%
}

\begin{document}

\maketitle
\thispagestyle{empty}
\pagestyle{empty}

\begin{abstract}

Robot assisted endoluminal intervention is an emerging technique for both benign and malignant luminal lesions. With vision-based navigation, when combined with pre-operative imaging data as priors, it is possible to recover position and pose of the endoscope without the need of additional sensors.
In practice, however, aligning pre-operative and intra-operative domains is complicated by significant texture differences.
Although methods such as style transfer can be used to address this issue, they require large datasets from both source and target domains with prolonged training times.
This paper proposes an efficient domain transfer method based on stylized Gaussian splatting, only requiring a few of real images (10 images) with very fast training time. 
Specifically, the transfer process includes two phases. In the first phase, the 3D models reconstructed from CT scans are represented as differential Gaussian point clouds. 
In the second phase, only color appearance related parameters are optimized to transfer the style and preserve the visual content. A novel structure consistency loss is applied to latent features and depth levels to enhance the stability of the transferred images. 
Detailed validation was performed to demonstrate the performance advantages of the proposed method compared to that of the current state-of-the-art, highlighting the potential for intra-operative surgical navigation.

\end{abstract}

\section{INTRODUCTION}

Recent advances in endoscopy have led to the development of endoluminal intervention. In these procedures, however, the narrow and tortuous luminal structures make it challenging to manipulate flexible endoscopes, necessitating efficient navigation algorithms. Accurate localization of the endoscope, a prerequisite for navigation, is hindered by complex artifacts and the absence of distinctive surface textures or anatomical landmarks.
A promising approach to address these challenges involves combining pre-operative CT scans with intra-operative endoscopic images, providing a robust structural prior for pose estimation \cite{shen2019context}. With this approach, multi-slice pre-operative CT is used to reconstruct a 3D virtual scene to provide both surgical planning and intra-operative guidance.
However, aligning these two imaging modalities is a challenge due to substantial texture differences between them.

To overcome this challenge, deep learning-based style transfer methods have made remarkable progresses in recent years. 
They capture the distributions of both domains at the latent features level and employ learning-based techniques -- such as Generative Adversarial Networks (GANs) \cite{eaGAN, jhu, cltsGAN, augment, cyclegan, cut, attentiongan, IBGAN, MI2GAN}, Energy-Based Models (EBMs) \cite{zhao2020learning,zhao2021unpaired,vqvae,tiwary2024cycle}, and diffusion-based models \cite{diffusion, unit-ddpm, EGSDE, unsb} -- to bridge appearance gap between the two domains.

\begin{figure}[!t]
    \centering
    \includegraphics[width=\linewidth]{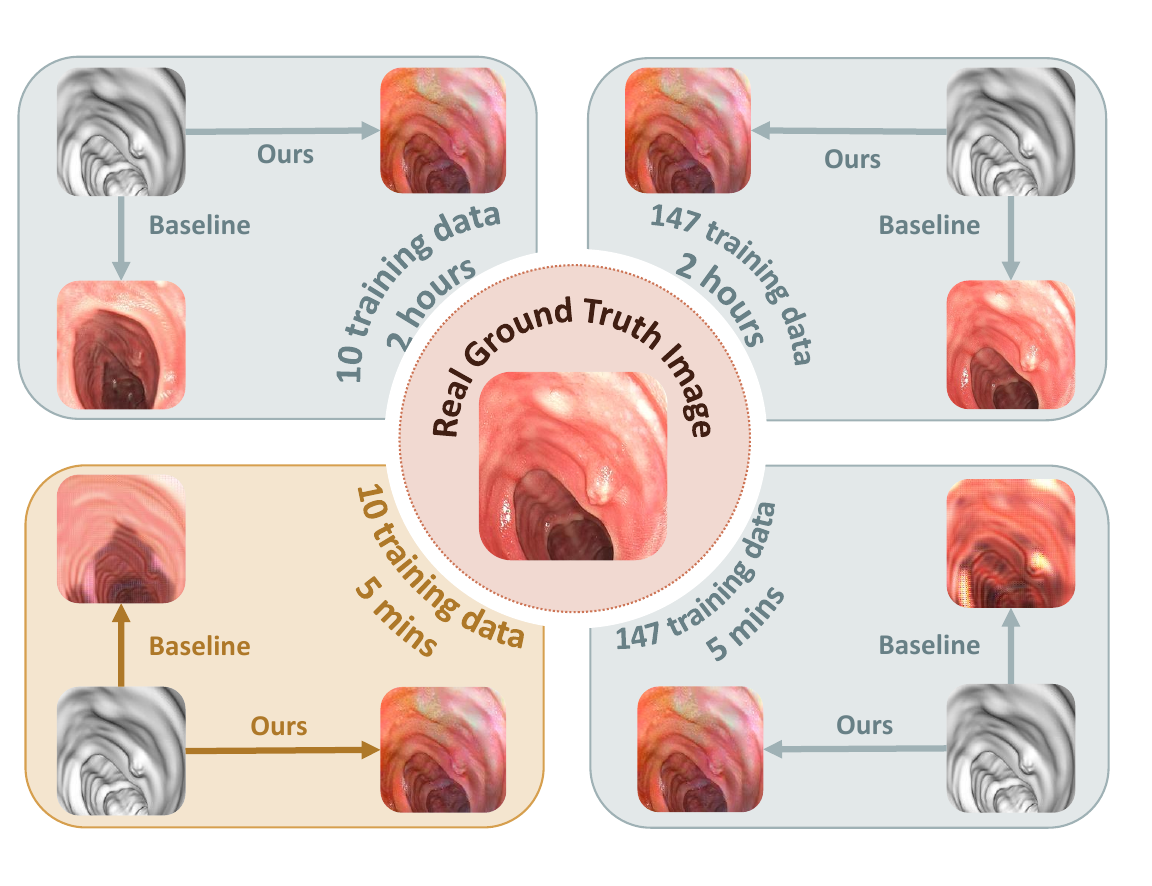}
    
    \caption{Challenges in efficient style transfer. Given sufficient training data and time, existing style transfer networks are capable of producing high-quality images. However, in the absence of adequate training data, the style transfer method may generate target-like textures, but introduces noticeable content distortion. Additionally, with insufficient training time, the style transfer method tends to produce blurry images. Our method is capable of achieving high-quality style transfer even with limited samples and very fast training time.}
    
    \label{fig:intro} 
    \end{figure}

GANs operate by designing a generator that creates images resembling the target domain, while a discriminator evaluates the similarity between the generated and target domains. During training, the generator improves until the discriminator can no longer reliably distinguish between real and generated images. Notable advancements in GANs include models like CycleGAN \cite{cyclegan}, which introduced cycle-consistency loss to enable image translation without paired data, and CUT \cite{cut}, which leveraged contrastive learning to align corresponding patches between source and target domains without requiring strict one-to-one mapping.
In contrast to GANs, EBMs learn an energy function that assigns lower energy values to plausible images and higher values to implausible ones, thereby modeling a probability distribution over the data. CF-EBM \cite{zhao2020learning} used Langevin Dynamics to guide source image features toward lower energy regions. LETIT \cite{zhao2021unpaired} can address the slow convergence issue of CF-EBM by utilizing the low-dimensional latent representations from VQ-VAE2 \cite{vqvae}. Furthermore, CCT-EBM \cite{tiwary2024cycle} improved structure consistency by employing a dual EBM architecture in the feature space.
Another approach based on diffusion-based models \cite{diffusion, unit-ddpm, EGSDE, unsb}, which refines noisy images progressively through reverse diffusion processes, also demonstrates remarkable success in style transfer.

Existing methods, however, suffer from two major issues: time and data inefficiency.
For time inefficiency, current approaches require training a neural network from scratch, which is highly time-consuming. Insufficient training time can lead to underfitting of the transfer network, resulting in blurry images as shown in Fig. \ref{fig:intro}.
In terms of data inefficiency, training a neural network typically requires a substantial amount of data. When limited real-life data are available, existing style transfer networks suffer from overfitting. As shown in the second quadrant of Fig. \ref{fig:intro}, for example, when the network is trained with only 10 real images, it overfits to both the style and content of those images, leading to distortion of the original content.

In real clinical scenarios, style transfer must be performed as quickly as possible using limited data. To address this challenge, this work focuses on enhancing the efficiency of style transfer by using stylized Gaussian splatting.
The core innovation lies in representing virtual 3D models as differentiable 3D Gaussian point clouds and transferring only color appearance related parameters during the style transfer process. This differential representation of the pre-operative 3D model provides a structural prior, while the optimization of color-related parameters enables the transfer to a target-like style without compromising the structural content.
In designing the loss functions, we introduce a novel structure consistency loss, which ensures alignment with the input image content at both the latent features and depth levels, thereby improving the stability of the generated images.
Detailed validation has been performed to assess the performance gain when compared to the current state-of-the-art.
Additionally, we demonstrate how the method can be used for camera pose estimation.

\section{Methods}

\begin{figure*}[]
\centering
\includegraphics[width=0.9\linewidth]{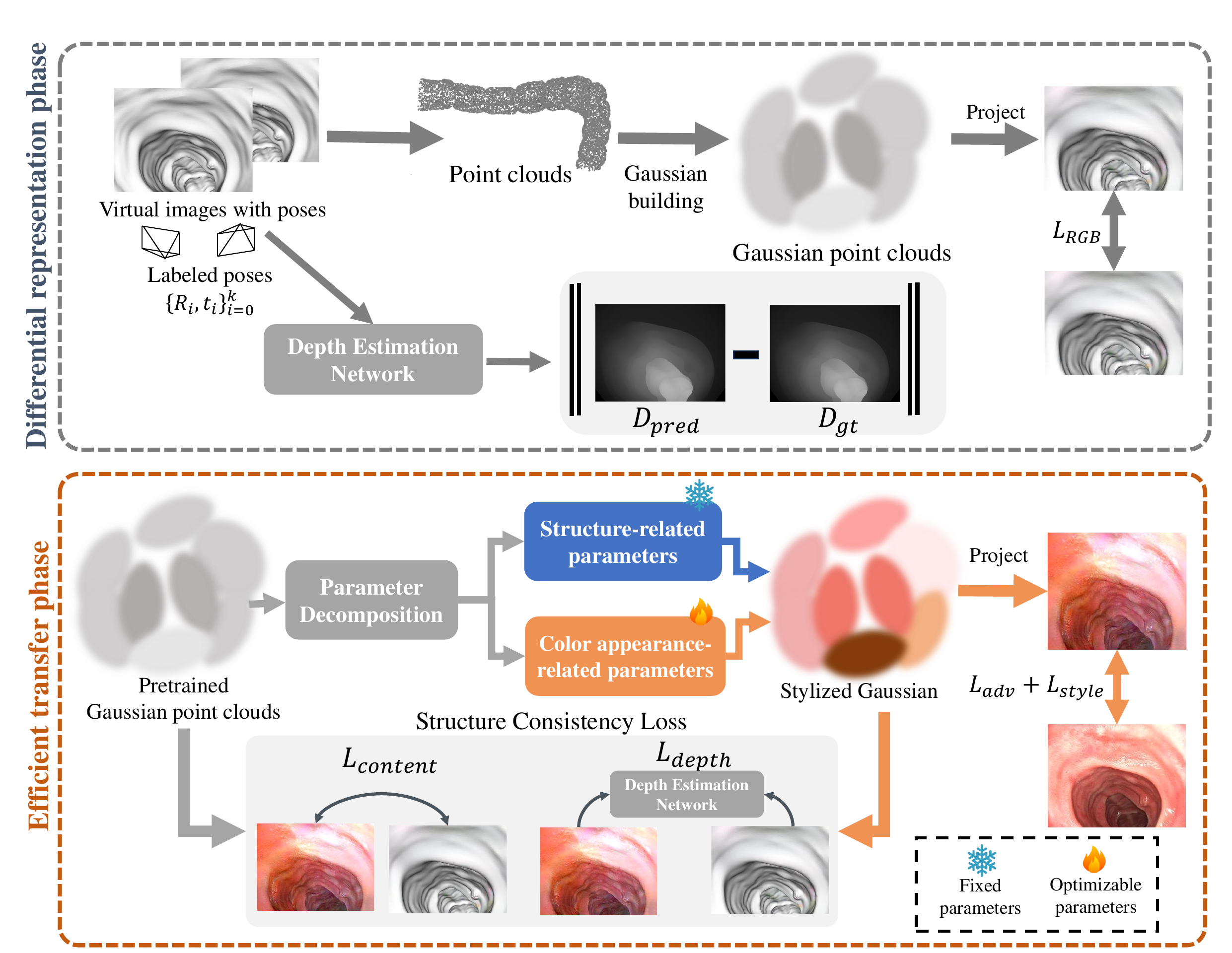}

\caption{The overall framework of our proposed method. In the differential representation phase, based on virtual images and poses, the Gaussian point clouds can be built based on Gaussian splatting theory. In the efficient transfer phase, the parameters of Gaussian point clouds are decomposed by structure-related parameters and color appearance-related parameters. Only color appearance-related parameters are optimized during this phase.}

\label{fig:framework} 
\end{figure*}

As illustrated in Fig. \ref{fig:framework}, the proposed framework consists of two phases: the differential representation phase and the efficient transfer phase.
In the differential representation phase, the 3D model reconstructed from CT scans is represented as Gaussian point clouds, which serve as a structural prior for the scene.
In the efficient transfer phase, we decompose the parameters in the pretrained Gaussian point clouds into structure-related and color appearance-related parameters. To preserve the original visual content while achieving target-like texture, only the color appearance-related parameters are optimized during the transfer process.
Additionally, we introduce a structure consistency loss function based on both latent features and depth information. This loss function ensures that the structural content remains consistent with input images throughout the transfer process.

\subsection{Differential representation phase}

For the pre-operative 3D models reconstructed from CT, existing methods typically represent them in non-differentiable formats, such as point clouds or meshes. However, these representations are inadequate for providing a structural prior during the style transfer process.
In contrast, this work introduces a differentiable representation of the 3D model based on Gaussian Splatting theory, which serves as a structural prior.

Gaussian splatting \cite{3dgs} is an explicit 3D scene representation based on a sparse set of SfM points. Given the 3D points, each Gaussian is defined by a 3D covariance matrix $\Sigma$ and a center point $x$:
\begin{equation}
    G(x) = e^{-\frac{1}{2}x^{T}\Sigma^{-1}x}
\end{equation}
where the covariance matrix $\Sigma$ can be represented by a rotation matrix $R$ and a scaling matrix $S$ for differentiable optimization:
\begin{equation}
    \Sigma = RSS^{T}R^{T}
\end{equation}

For projecting 3D Gaussian points to 2D images,
the rendering process of N ordered points overlapping a pixel follows a specific formula:
\begin{equation}
    C = \sum_{i \in N} c_{i}\alpha_{i}\prod_{j=1}^{i-1}(1-\alpha_{j})
\end{equation}
where $c_{i}, \alpha_{i}$ indicate the color and density of a given point. These parameters are determined by a Gaussian with a covariance matrix $\Sigma$, which is then scaled by optimizable per-point opacity and spherical harmonics color coefficients.

In the differential representation phase, a pre-operative 3D model can be traversed to generate numerous virtual images with corresponding camera poses. According to the outlined formulations, a 3D scene can be represented by multiple Gaussian points. Each Gaussian point is characterized by the following optimizable attributes: position x$\in \mathbb{R}^{3}$, spherical harmonics coefficients c$\in \mathbb{R}^{k}$ (where k indicates the degree of freedom), opacity $\alpha \in \mathbb{R}$, rotation quaternion $q \in \mathbb{R}^{4}$, and a scaling factor $s \in \mathbb{R}^{3}$. 
The mean squared error (MSE) is used to optimize all parameters:
\begin{equation}
    L_{RGB} = \sum_{r \in R} \|C_{r} - \hat{C}_{r}\|^{2}
\end{equation}
where R is the set of rays in each training batch, $C_{r}$ and $\hat{C}_{r}$ are the predicted and ground truth color. 

At the same time, using images and corresponding depth ground truth from the virtual domain, our method trains a depth estimation network, $D_{v}$. $D_{v}$ has the same architecture with \cite{depthnet}, optimized by the MSE loss function.

\subsection{Efficient transfer phase}

In this phase, we apply style transfer to the pretrained 3D Gaussian point clouds, generating target-like images while preserving the input visual content.
Given the pretrained 3D Gaussian point clouds $\Theta$, where each point $\Theta_{i} = \{ x_{i}, s_{i}, q_{i}, \alpha_{i}, c_{i} \}$ represents the parameters of the $i$-th Gaussian point, this phase aims to obtain a stylized 3D Gaussians $\Theta^{r}$.
$\Theta^{r}$ represents a 3D model that aligns with the style of the real domain, while faithfully preserving the original structural content of the virtual model.

As illustrated in Fig. \ref{fig:framework}, the parameters of 3D Gaussian point clouds are categorized into two groups: structure-related parameters (position, opacity, rotation quaternion, and scaling factor) and color appearance-related parameters (spherical harmonics coefficients). By fixing the structure-related parameters, each instance of Gaussian point clouds consistently represents a specific spatial structure, regardless of domain texture variations. Our framework optimizes only the color appearance-related parameters during the efficient transfer phase to effectively transfer style while preserving the original visual content.

Additionally, we introduce two categories of loss functions: style transfer loss and structure consistency loss. 
For the style transfer loss, we incorporate a \textit{global loss} that captures the overall style of the target domain and a \textit{local loss} that emphasizes local features. To ensure the preservation of the visual content, we propose a novel \textit{structure consistency loss} at both the latent features and depth levels.

\paragraph{Global loss} This loss function aims to ensure that the generated image closely resembles the target image in overall style by focusing on the statistical similarity of latent features. 
Specifically, it minimizes the mean squared error in feature statistics between the generated image $I_{g}$ and the real image $I_{r}$ as follows:
\begin{eqnarray}
\begin{aligned}
    L_{style} = \sum_{l}||\mu(\tau_{l}(I_{g}))  - \mu(\tau_{l}(I_{r})) ||_{2}\\
    + \sum_{l}||\Sigma(\tau_{l}(I_{g}))  - \Sigma(\tau_{l}(I_{r})) ||_{2}
\end{aligned}
\end{eqnarray}
where $\mu$ and $\Sigma$ represent the mean and standard deviation respectively, and $\tau_{l}()$ denotes the feature representation obtained from the $l$-th layer of VGG-19 \cite{vgg}.

\paragraph{Local loss} Different from the global loss, local loss focuses on local texture features. Based on the local properties of the convolution operation, we incorporate an adversarial training loss:
\begin{eqnarray}
\begin{aligned}
    L_{D} &= \mathbb{E}_{r \sim p(I_{r})}[\log(D(r))] + \mathbb{E}_{g \sim p(I_{g})}[1 - \log(D(g))] \\
    L_{adv} &= \mathbb{E}_{g \sim p(I_{g})}[\log(D(g))]
\end{aligned}
\end{eqnarray}
where $D$ is a discriminator from the standard GAN \cite{goodfellow2014generative} to distinguish conditions from each domain.

\paragraph{Structure consistency loss} In addition to transferring the style to the target domain, it is crucial to preserve the visual content of the input image. Therefore, this work introduces a novel structure consistency loss in both latent features and depth levels.
For the latent features level, $L_{content}$ ensures that the generated image $I_{g}$ and the virtual image $I_{v}$ exhibit similar content features:
\begin{equation}
    L_{content} = ||\tau(I_{g}) - \tau(I_{v})||_{2}
\end{equation}
where $\tau()$ is the feature representation obtained from the $relu4\_1$ layer of a pretrained VGG-19 network.

At the depth level, since depth estimation networks emphasize structural information over texture details, the depth maps of input virtual image $I_{v}$ and the generated image $I_{g}$ should be consistent. Therefore, 
we use depth similarity as a constraint to enhance structure consistency:
\begin{equation}
    L_{depth} = ||D_{v}(I_{v}) - D_{v}(I_{g})||_{2} + \sum_{i=1}^{K}||E_{i}(I_{v}) - E_{i}(I_{g})||_{2}
\end{equation}
where $E_{i}$ is the $i$-th encoder layer of the pretrained depth network $D_{v}$, and $K$ is the number of layers of encoder $E$.

Overall, the all loss function in efficient transfer phase is: 
\begin{equation}
    L_{efficient} = L_{style} + L_{adv} + L_{content} + L_{depth}
\end{equation}

\begin{figure*}[]
\centering
\includegraphics[width=0.9\linewidth]{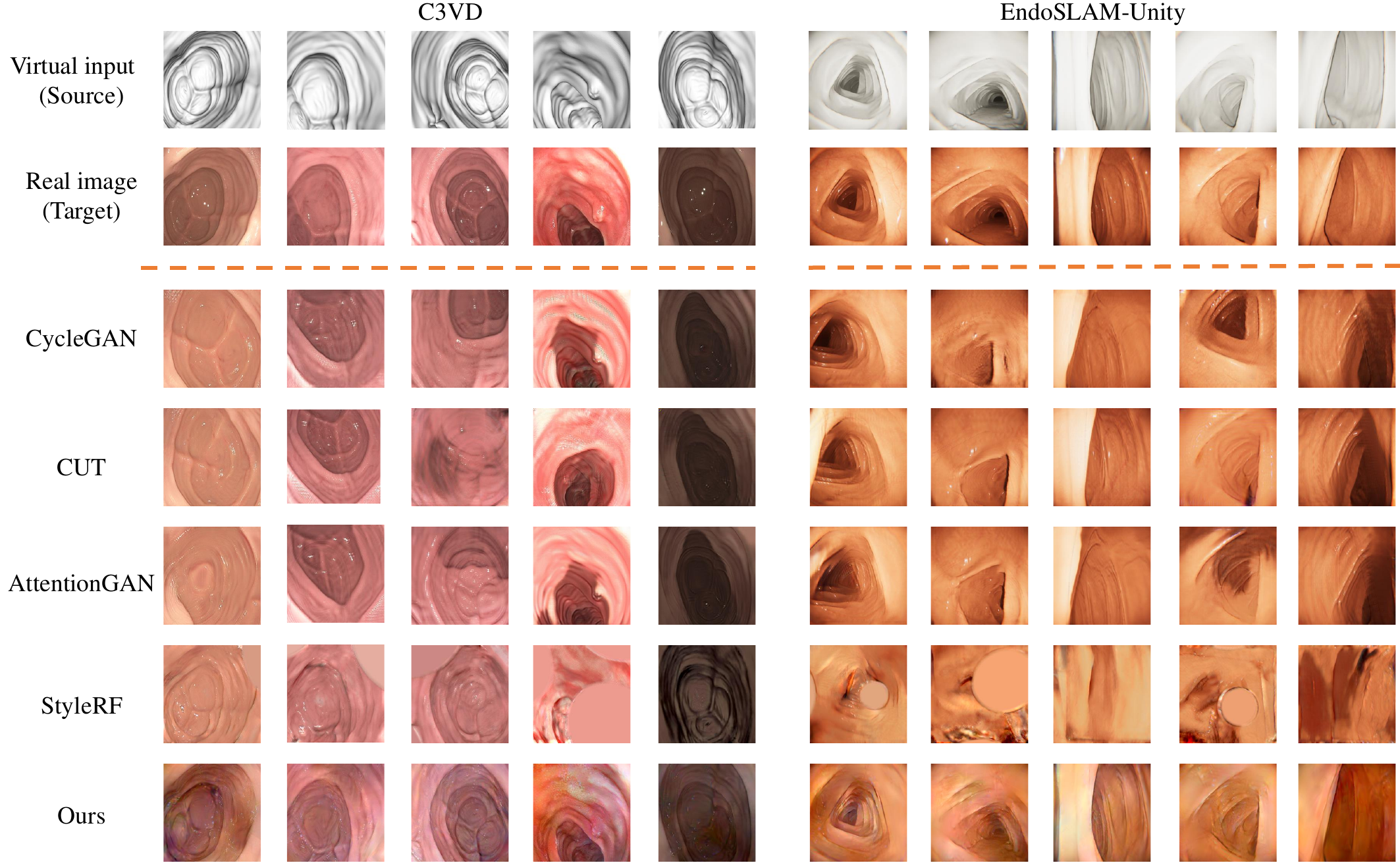}
\caption{Examples of the virtual inputs, target real images, and the generated results using different methods. It is evident that GAN-based methods exhibited content bias, StyleRF generated blurry images, while our method effectively transferred the overall style and preserved the input structural content.}

\label{fig:style_vis} 
\end{figure*}

\begin{table*}[t!]
\centering
\caption{Quantitative results of FID, KID, SSIM and PSNR. The best results are in \textbf{bold}.}
\resizebox{0.95\linewidth}{!}{
\begin{tabular}{cccccccccccc}
\hline
   \multirow{2}{*}{Method}    & \multirow{2}{*}{Transfer Time}   & \multirow{2}{*}{FPS}   & \multicolumn{4}{c}{C3VD} &  & \multicolumn{4}{c}{EndoSLAM-Unity}  \\ \cline{4-7} \cline{9-12} 
      &    &   & FID  & KID & SSIM & PSNR &  & FID    & KID    & SSIM    & PSNR    \\ \hline
CycleGAN  &50 mins & 39.54  &  205.174    &   0.272  &  0.425    &   12.461   &  &  147.976      & 0.170       &   0.552     &   16.005    \\ 
CUT   &100 mins   & 111.32   &   314.612   &  0.446    &  0.400    &  10.286    &  &   179.799     & 0.216       &   0.572      &    16.371      \\ 
AttentionGAN  &70 mins & 32.34 &  405.559    &  0.585   & 0.338     &  11.955    &  &   187.237     & 0.206      &   0.588      &    17.130       \\ 
StyleRF     &5 hours &  6.23  &  195.746    &  0.214   &  0.488    &  14.286    &  &    326.469    & 0.522       &  0.521       &    14.300       \\ 
Ours      &5 mins & 126.03   & \textbf{168.952}     &  \textbf{0.173}   &  \textbf{0.565}    &  \textbf{16.472}    &  &  \textbf{132.519}      &  \textbf{0.153}      &    \textbf{0.599}     &    \textbf{18.735}    \\ \hline
\end{tabular}
}
\label{tab:synthetic}
\end{table*}

\begin{table*}[]
\centering
\caption{Quantitative results of baselines and proposed method. The best results are in \textbf{bold}.}
\resizebox{0.95\linewidth}{!}{
\begin{tabular}{ccccccccc}
\hline
\multirow{2}{*}{Method}  & \multicolumn{3}{c}{C3VD}        &                                                 & \multicolumn{3}{c}{EndoSLAM-Unity}    \\  \cline{2-4} \cline{6-8} 

                      &\multicolumn{1}{c}{ATE(mm)} & \multicolumn{1}{c}{$t_{rpe}$(mm)} & \multicolumn{1}{c}{$R_{rpe}$(deg)} & &  \multicolumn{1}{c}{ATE(mm)} & \multicolumn{1}{c}{$t_{rpe}$(mm)} & \multicolumn{1}{c}{$R_{rpe}$(deg)}  \\ \hline

 Virtual only    &\multicolumn{1}{c}{10.381$\pm$5.229}        & \multicolumn{1}{c}{0.635$\pm$0.374}       & \multicolumn{1}{c}{0.827$\pm$0.298}   &        & \multicolumn{1}{c}{9.453$\pm$4.082}        & \multicolumn{1}{c}{0.639$\pm$0.291}       & \multicolumn{1}{c}{1.021$\pm$0.380}      \\

CycleGAN     &  \multicolumn{1}{c}{8.085$\pm$5.154}        & \multicolumn{1}{c}{0.366$\pm$0.242}       & \multicolumn{1}{c}{0.411$\pm$0.474}   &        & \multicolumn{1}{c}{5.298$\pm$3.568}        & \multicolumn{1}{c}{0.269$\pm$0.140}       & \multicolumn{1}{c}{1.136$\pm$0.432}      \\ 
CUT    &  \multicolumn{1}{c}{7.048$\pm$2.762}        & \multicolumn{1}{c}{0.425$\pm$0.249}       & \multicolumn{1}{c}{0.406$\pm$0.352}    &       & \multicolumn{1}{c}{7.337$\pm$1.522}        & \multicolumn{1}{c}{0.497$\pm$0.362}       & \multicolumn{1}{c}{0.933$\pm$0.270}         \\
AttentionGAN      &  \multicolumn{1}{c}{6.891$\pm$3.472}        & \multicolumn{1}{c}{0.328$\pm$0.165}       & \multicolumn{1}{c}{0.346$\pm$0.344}    &      & \multicolumn{1}{c}{5.178$\pm$3.373}        & \multicolumn{1}{c}{0.279$\pm$0.146}       & \multicolumn{1}{c}{0.871$\pm$0.288}      \\
StyleRF    & \multicolumn{1}{c}{7.016$\pm$2.423}        & \multicolumn{1}{c}{0.430$\pm$0.266}       & \multicolumn{1}{c}{0.532$\pm$0.772}    &       & \multicolumn{1}{c}{6.819$\pm$2.058}        & \multicolumn{1}{c}{0.545$\pm$0.287}       & \multicolumn{1}{c}{1.726$\pm$0.729}     \\ 
Ours    & \multicolumn{1}{c}{\textbf{3.761$\pm$2.684}}        & \multicolumn{1}{c}{\textbf{0.189$\pm$0.087}}       & \multicolumn{1}{c}{\textbf{0.283$\pm$0.353}}  &        & \multicolumn{1}{c}{\textbf{4.973$\pm$1.761}}        & \multicolumn{1}{c}{\textbf{0.162$\pm$0.028}}       & \multicolumn{1}{c}{\textbf{0.887$\pm$0.130}}    
\\ \hline
\end{tabular}
}
\label{tab:pose}
\end{table*}

\section{Experiment}

\subsection{Datasets and evaluation} We conducted our experiments using two publicly available datasets: C3VD \cite{c3vd} and EndoSLAM-Unity \cite{endoslam}.
The C3VD dataset contains 22 video sequences captured from real colonoscopies, each paired with corresponding virtual models. 
We imported the 3D virtual models and render the virtual images by traversing the entire 3D virtual models. 
For each video sequence, all virtual frames were utilized to reconstruct the virtual 3D model, while only the first 10 real frames were employed for style transfer training. The remaining real data were used for testing.
EndoSLAM-Unity offers a Unity environment with virtual and real textures. We sampled 5868 paired images and split them into 10 videos for training and validation. Similar to the C3VD dataset, for each video sequence, all virtual frames were used to reconstruct the virtual 3D model, and only the first 10 real frames were used for training. The remaining real data were used for testing.
Our framework was evaluated against various domain adaptation techniques, including CycleGAN \cite{cyclegan}, CUT \cite{cut}, AttentionGAN \cite{attentiongan}, and StyleRF \cite{stylerf}. CycleGAN, CUT, and AttentionGAN utilized generative adversarial networks to transfer the source domain to the target domain. In contrast, StyleRF first trained a NeRF model and then transformed the grid features according to the reference style.
To ensure a fair comparison of training time, we recorded the training time for all methods until convergence.

To comprehensively validate our experiments, we used traditional unpaired image translation metrics: Fréchet Inception Distance (FID) \cite{fid} and Kernel Inception Distance (KID) \cite{kid}. We also measured Structural Similarity Index (SSIM) and Peak Signal-to-Noise Ratio (PSNR) for precise evaluation.
Additionally, we demonstrated that our method can be used for downstream sim-to-real tasks for camera pose estimation.
Specifically, we applied style transfer methods to virtual images to generate real-like images with corresponding pose ground truth. These generated images, along with their ground truth poses, were then used to train an absolute pose estimation network \cite{kendall2015posenet}, which we subsequently tested in the real domain. Average Trajectory Error (ATE), $t_{rpe}$, and $R_{rpe}$ were used as the evaluation metric.

\subsection{Implementation details}

During the training phase, all models were optimized using the Adam optimizer. For the differential representation phase, we trained with a feature learning rate of 0.0025.
At the same time, we trained the depth network using virtual images and depth ground truth.
As the representation phase occurred pre-operatively, the training was conducted offline. For the efficient transfer phase, we increased the feature learning rate to 0.025 and optimized color appearance-related parameters. Code is at: \href{https://github.com/EndoluminalSurgicalVision-IMR/Sim2Real5Mins}{https://github.com/EndoluminalSurgicalVision-IMR/Sim2Real5Mins}.

\begin{figure*}[]
\centering
\includegraphics[width=0.9\linewidth]{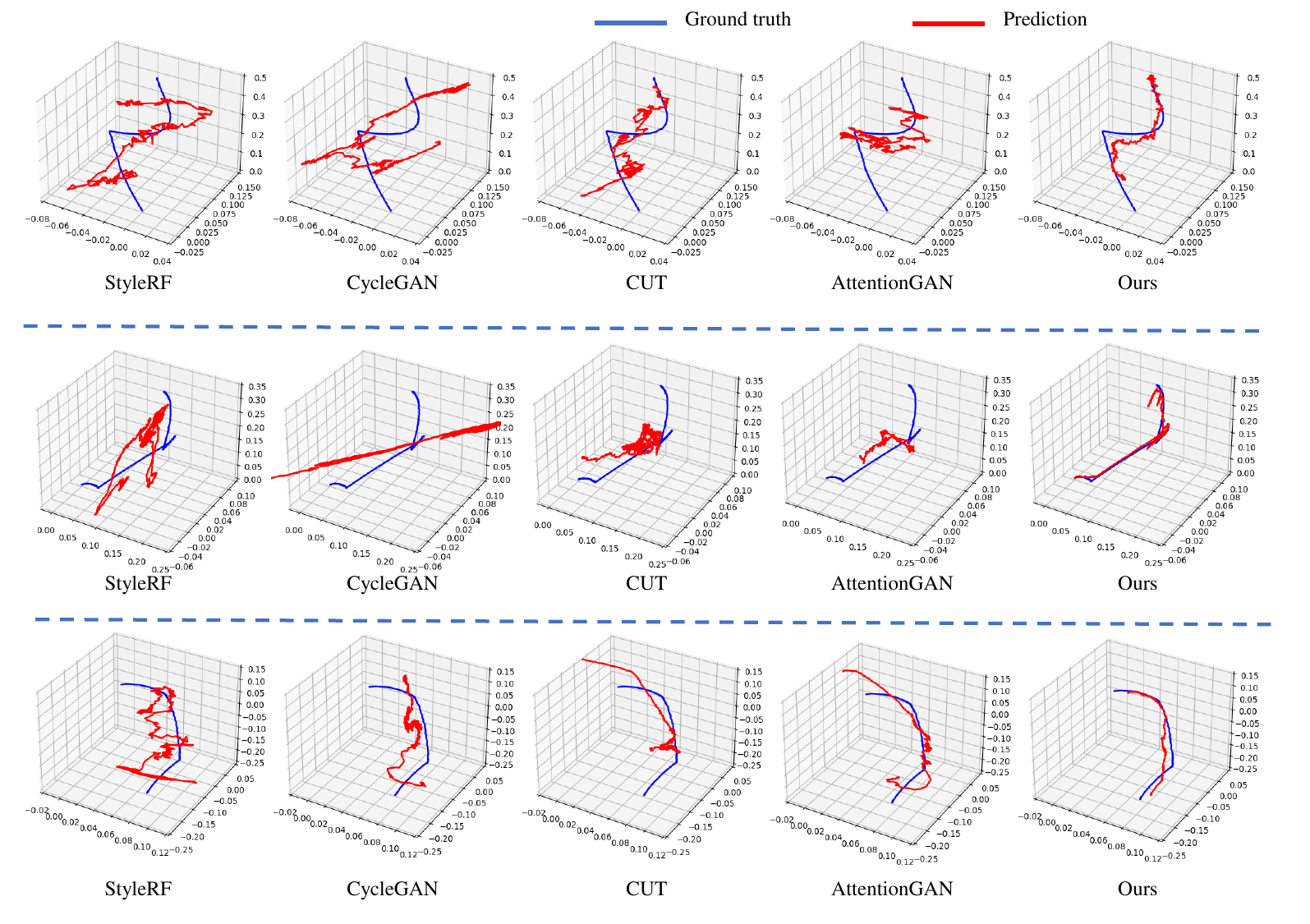}

\caption{Qualitative results on three sequences. Due to the lack of training images, the images generated by the baselines exhibit structural deviations, resulting in suboptimal performance in pose estimation. Our method preserves the structural content, achieving superior results in pose estimation.}

\label{fig:pose_vis} 
\end{figure*}

\subsection{Results}

\subsubsection{Image translation results}
Table \ref{tab:synthetic} presents the quantitative comparison, demonstrating that our framework significantly outperform other methods across all metrics. Additionally, t-tests were conducted to compare different methods. The p-values of PSNR between our method and CycleGAN were 6,8e-10, 3.1e-7 on two datasets, indicating statistically significant improvements.
The ``Transfer time'' column reports the time required for training until convergence. GAN-based methods, which require training a neural network from scratch, are highly time-intensive. Similarly, StyleRF’s transformation of grid features is also computationally expensive. In contrast, our method utilizes structural information from pre-operative Gaussian point clouds, requiring only the optimization of color appearance-related parameters, enabling completion within just five minutes.
Furthermore, our method achieves the highest test-time FPS, demonstrating superior efficiency compared to other approaches.

In Fig. \ref{fig:style_vis}, we visualized the results of the style transfer. Although GAN-based approaches successfully emulated the style of the real domain, they introduced noticeable content distortion. This issue arose from the limited number of unpaired real images used during training, which caused the style transfer network to overfit both the style and structural content of the unpaired training images, resulting in generated images that deviated from the original visual content. The NeRF-based method, StyleRF, led to content blurring.
In contrast, our method was capable of generating target-like images while preserving the input visual content, even with minimal training time.

\subsubsection{Pose estimation results}
To validate the effectiveness of style transfer in endoscopic navigation, we employed pose estimation as a supplementary task to verify the alignment of virtual and real domain.
The accuracy of pose estimation is presented in Table \ref{tab:pose} and Fig. \ref{fig:pose_vis}. 
The \textit{virtual only} setting demonstrated suboptimal performance owing to the substantial appearance gap between virtual and real domains. 
For baseline methods, using only 10 real images for training introduced content bias in the generated images, resulting in failures in the pose estimation task. In contrast, our method, which optimized only color appearance-related parameters, successfully generated target-like images while preserving the original visual content, achieving superior performance.

\begin{table}[]
\centering
\caption{Ablation results of proposed modules on the C3VD dataset.}
\resizebox{\linewidth}{!}{
\begin{tabular}{cccc}
\hline
\multirow{2}{*}{Method} & \multicolumn{3}{c}{C3VD}                                                                                                                                \\ \cline{2-4} 
                        & \multicolumn{1}{c}{ATE(mm)} & \multicolumn{1}{c}{$t_{rpe}$(mm)} & \multicolumn{1}{c}{$R_{rpe}$(deg)}  \\ \hline
w/o global loss                & \multicolumn{1}{c}{4.458$\pm$2.232}        & \multicolumn{1}{c}{0.279$\pm$0.096}       & \multicolumn{1}{c}{0.313$\pm$0.162}         \\
w/o local loss                     & \multicolumn{1}{c}{4.810$\pm$2.129}        & \multicolumn{1}{c}{0.293$\pm$0.107}       & \multicolumn{1}{c}{0.328$\pm$0.140}          \\
w/o structure consistency loss            & \multicolumn{1}{c}{6.850$\pm$2.865}        & \multicolumn{1}{c}{0.314$\pm$0.242}       & \multicolumn{1}{c}{0.349$\pm$0.195}     \\ 
Ours                   & \multicolumn{1}{c}{3.761$\pm$2.684}        & \multicolumn{1}{c}{0.189$\pm$0.087}       & \multicolumn{1}{c}{0.283$\pm$0.153}         
\\ \hline
\end{tabular}
}
\label{tab:ablation}
\end{table}

\section{Discussion}

\subsection{Ablation study}

\begin{figure}[]
\centering
\includegraphics[width=\linewidth]{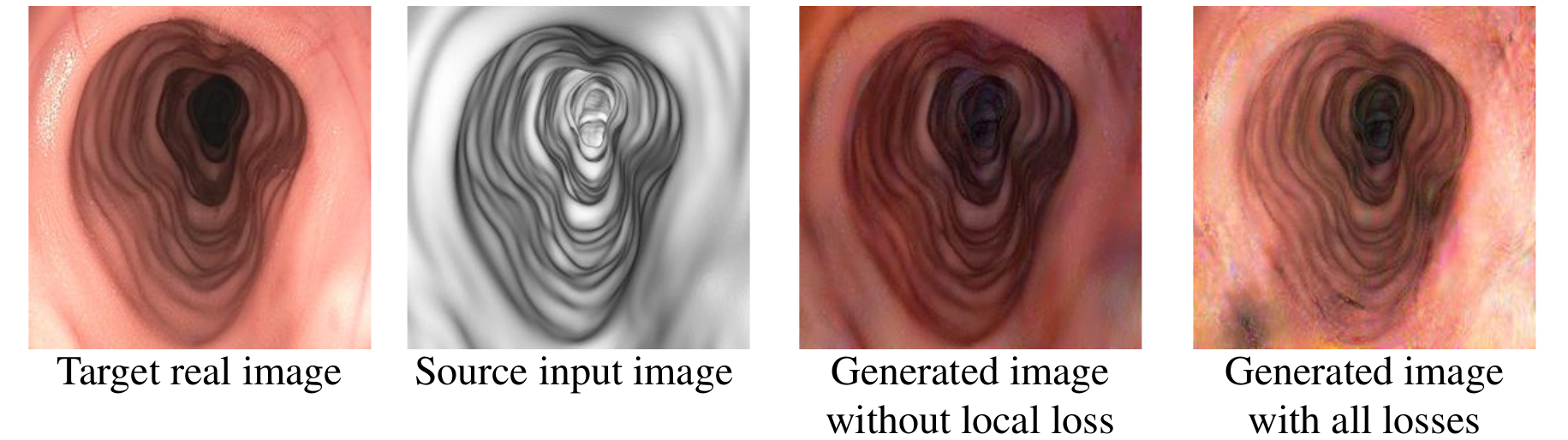}

\caption{Visual analysis of the effects of the global loss and the local loss.}

\label{fig:ablation} 
\end{figure}

We conducted ablation studies on the C3VD dataset to investigate the significance of each loss function. Table \ref{tab:ablation} presents the results, demonstrating the considerable importance of each loss function. Notably, removing the structure consistency loss deteriorated pose estimation performance, as maintaining input structure consistency is crucial for camera pose estimation.
Furthermore, we performed a visual analysis of the effects of global and local loss. As shown in Fig. \ref{fig:ablation}, using only global loss led to generated images that matched the overall style of the target but exhibited deviations in local details. By incorporating the local loss, the generated images captured local features more precisely, resulting in a closer resemblance to the target image.

\subsection{Limitation}
Although our work achieves efficient style transfer with minimal data and training time, it relies on the modeling capabilities of Gaussian Splatting. For larger clinical scenes, such as bronchial trees, Gaussian Splatting struggles to accurately reconstruct the structure, hindering effective style transfer. A promising direction for addressing this limitation is to model large-scale scenes using multiple localized Gaussian Splatting representations.
Another issue is the non-rigid deformations, a direct mapping between preoperative CT and intraoperative endoscopic views may not always be valid. 
A viable approach to addressing this limitation could involve integrating a 4D Gaussian Splatting framework to dynamically account for temporal and spatial deformations. 

\section{Conclusion}

In this paper, we introduce an efficient domain transfer method based on stylized Gaussian splatting. By using Gaussian splatting as the structural prior and optimizing only color appearance-related parameters, our approach can generate target-like images with minimal real images and training time. Additionally, we design a novel structure consistency loss to preserve the visual content. Comprehensive experiments demonstrate that our method achieves state-of-the-art performance in image translation and camera pose estimation tasks, with just five minutes of training.

\bibliographystyle{IEEEtran}
\bibliography{IEEEabrv,refs_icra}

\end{document}